\pdfoutput=1

\documentclass[11pt]{article}

\usepackage[final]{acl}

\usepackage{times}
\usepackage{latexsym}

\usepackage[T1]{fontenc}

\usepackage[utf8]{inputenc}

\usepackage{enumitem}

\usepackage{microtype}
\usepackage{tikz}
\usepackage{booktabs}
\usepackage[table]{xcolor}
\usepackage{multicol,multirow}
\usepackage{makecell}
\usepackage[edges]{forest}
\usepackage{pgfplots}
\pgfplotsset{compat=1.18}
\usepackage{hyperref}
\usepackage[framemethod=tikz]{mdframed}
\usepackage{subcaption}
\usepackage{amsmath}
\definecolor{lgreen}{rgb}{0.89,0.94,0.85}
\definecolor{lred}{rgb}{0.98, 0.90, 0.84}
\definecolor{lyellow}{rgb}{1.00, 0.95, 0.80}
\definecolor{lblue}{rgb}{0.85, 0.89, 0.95}
\definecolor{hidden-draw}{RGB}{20,68,106}
\definecolor{hidden-pink}{RGB}{255,245,247}
\definecolor{bblue}{rgb}{0.95,0.97,0.98}
\definecolor{blueref}{rgb}{0.005,0.0,0.478}
\definecolor{la}{RGB}{128,208,195}
\definecolor{lb}{RGB}{158,170,196}
\definecolor{lc}{RGB}{128,208,195}
\definecolor{ld}{RGB}{158,170,196}
\definecolor{le}{RGB}{128,208,195}
\definecolor{bk}{RGB}{158,170,196}

\newcommand{\revised}[1]{\textcolor{black}{#1}}
\newcommand{\needrevise}[1]{\textcolor{red}{#1}}

\tikzset{%
    parent/.style =          {align=center,text width=0.7cm, rounded corners=2pt, line width=0.8mm, fill=white!0, draw=white!90},
    child/.style =           {align=center,text width=1.4cm,rounded corners=2pt, fill=blue!10,draw=blue!90,line width=0.3mm},
    T1/.style =           {align=center,text width=1.8cm,rounded corners=3pt, fill=lblue!100, draw=black,line width=0.2mm},   
    T1_end/.style =           {align=left, text width=5cm,rounded corners=5pt, fill=lblue!100,draw=black!0,line width=0.3mm},
    T2/.style =           {align=center,text width=1.8cm,rounded corners=3pt, fill=lred!100, draw=black,line width=0.2mm},   
    T2_end/.style =           {align=left, text width=5cm,rounded corners=5pt, fill=lred!100,draw=black!0,line width=0.3mm},
    T3/.style =           {align=center,text width=1.8cm,rounded corners=3pt, fill=lyellow!100, draw=black,line width=0.2mm},   
    T3_end/.style =           {align=left, text width=5cm,rounded corners=5pt, fill=lyellow!100,draw=black!0,line width=0.3mm},
    T4/.style =           {align=center,text width=1.8cm,rounded corners=3pt, fill=lgreen!100, draw=black,line width=0.2mm},   
    T4_end/.style =           {align=left, text width=5cm,rounded corners=5pt, fill=lgreen!100,draw=black!0,line width=0.3mm}
}

\usepackage{inconsolata}

\usepackage{graphicx}

\title{When Large Language Models Meet Speech:\\A Survey on Integration Approaches}

\author{
  \revised{Zhengdong Yang, Shuichiro Shimizu, Yahan Yu, Chenhui Chu}\\
  \revised{Kyoto University}\\
  \revised{\texttt{\{zd-yang, sshimizu, yahan\}@nlp.ist.i.kyoto-u.ac.jp} \quad \texttt{chu@i.kyoto-u.ac.jp}}
}

\begin{document}
\maketitle
\begin{abstract}
Recent advancements in large language models (LLMs) have spurred interest in expanding their application beyond text-based tasks. A large number of studies have explored integrating other modalities with LLMs, notably speech modality, which is naturally related to text.
This paper surveys the integration of speech with LLMs, categorizing the methodologies into three primary approaches: text-based, latent-representation-based, and audio-token-based integration.
We also demonstrate how these methods are applied across various speech-related applications and highlight the challenges in this field to offer inspiration for future research.
\end{abstract}

\section{Introduction}

In recent years, the field of natural language processing (NLP) has been greatly reshaped by the development of large language models (LLMs)~\cite{brown-etal-2020,touvron-etal-2023-llama,geminiteam-2024-gemini,bai-2023-qwen,deepseekai-2024-deepseek}.
These models have not only shown excellent ability in understanding and generating text but have also sparked interest in their potential applicability across other modalities, including speech.
The integration of speech and LLMs offers a wide range of potential applications, including speech translation, conversational chatbots, and enhanced human-computer interaction in robotics.


Survey papers have reviewed speech language models \citep{peng-etal-2024-speechllmsurvey,cui-etal-2025-speechlmsurvey}, as well as audio language models~\cite{latif-2023-large-audio} and multimodal language models~\cite{ghosh-2024-vision-language-model,zhang-2024-mmllm}.
However, there still lacks a survey specifically on the integration approaches of speech and LLMs, posing a challenge for researchers seeking to address this complex problem.

Distinct from other survey papers regarding speech and LLMs, this paper provides insight into the problem by specifically surveying the integration approaches of speech and LLMs.
Studies on LLM tokenization~\citep{chai-etal-2024-tokenization,tao-etal-2024-scalingvocab} suggests that tokenization methods can affect the performance of LLMs.
On the other hand, studies on tokenization methods for speech language modeling~\citep{gat-etal-2023-discreteslm,borsos-etal-2023-audiolm} also show that speech tokenization methods can affect the performance of speech language models.
In contrast to text processing, speech--LLM integration approaches are not limited to discrete tokenization, as presented in this paper.
Therefore, studying the integration between speech and LLMs can be a key to innovations.


\begin{figure}[t]
  \centering
  \includegraphics[width=\linewidth]{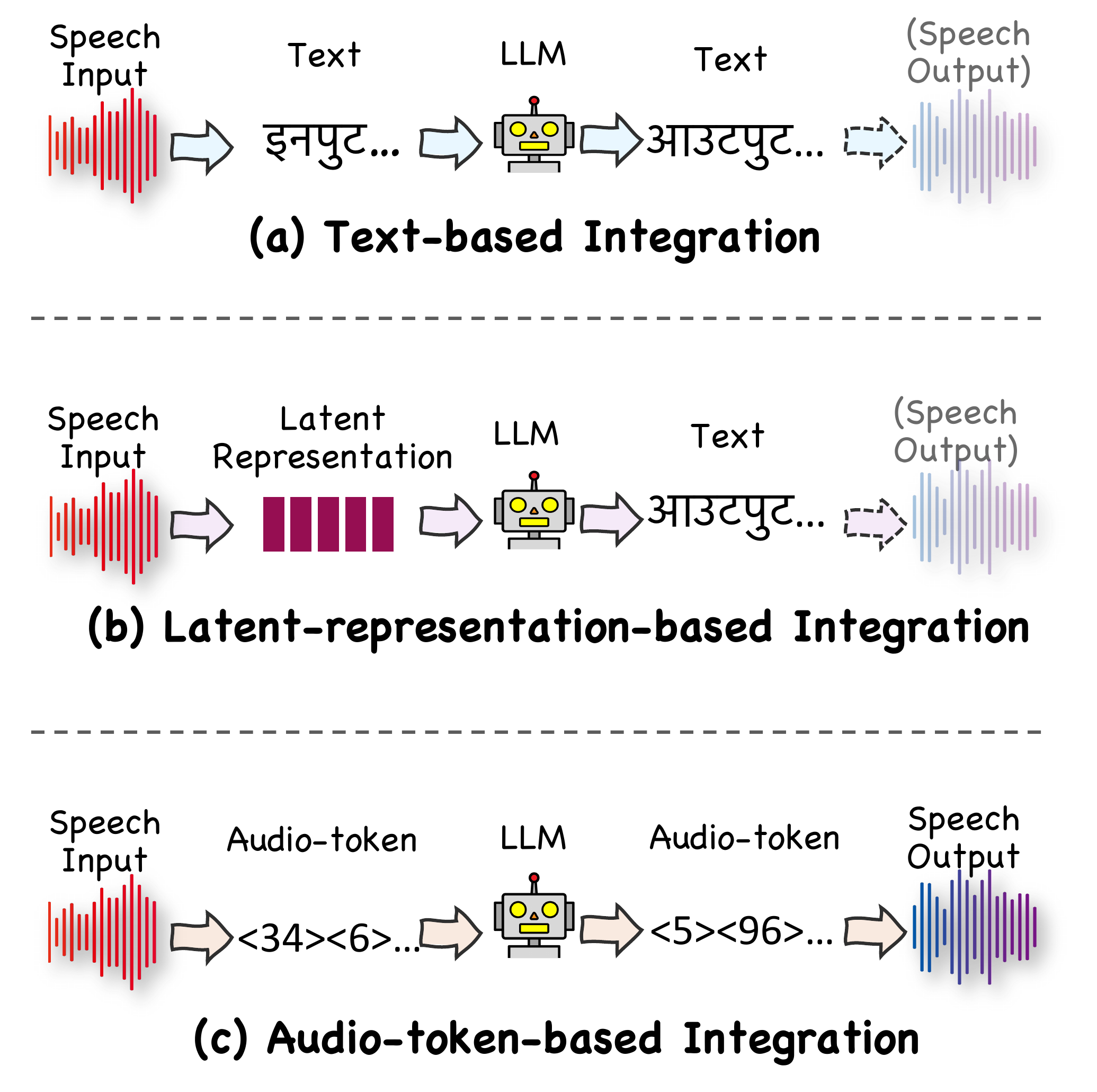}
  \caption{Overview of three fundamental approaches for speech--LLM integration.}
  \label{fig:overview}
\end{figure}

In this paper, we systematically categorize a substantial body of research on speech--LLM integration,\footnote{One challenge that affects the scope of studies is the lack of standard definition for LLMs. In this paper, we adopt the loose definition by \citet{zhao-etal-2023-survey-llm}, focusing on models with over 10 billion parameters, while also including notable studies with smaller models.}
and provide a clear taxonomy of the integration approaches.
We broadly categorize the integration into the following three types:
(a) \textbf{Text-based integration}: LLMs process textual data, integrated with speech-to-text and/or text-to-speech models;
(b) \textbf{Latent-representation-based integration}: Latent vector representations that encode speech data are utilized, mainly as inputs to LLMs;
(c) \textbf{Audio-token-based integration}: Speech tokens, such as semantic tokens and/or acoustic tokens, are used as the inputs/outputs for LLMs.
The overview of these approaches is illustrated in Figure \ref{fig:overview}, and the detailed taxonomy with representative studies is presented in Figure \ref{fig:taxonomy}.

\tikzstyle{my-box}=[
    rectangle,
    draw=hidden-draw,
    rounded corners,
    text opacity=1,
    minimum height=1.5em,
    minimum width=5em,
    inner sep=2pt,
    align=center,
    fill opacity=.5,
    line width=0.8pt,
]

\tikzstyle{leaf}=[my-box, minimum height=1.5em,
    fill=bblue!40, text=black, align=left,font=\normalsize,
    inner xsep=2pt,
    inner ysep=4pt,
    line width=0.8pt,
    draw=black,
]

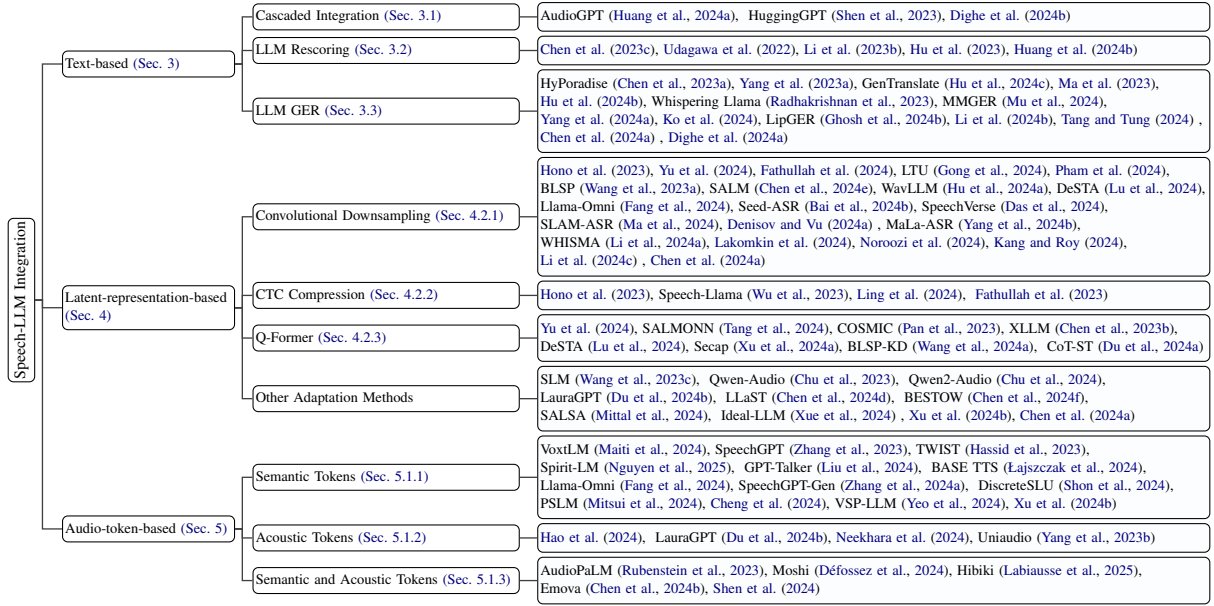
\begin{figure*}[t!]
    \centering
    \resizebox{1.0\textwidth}{!}{
        \begin{forest}
            forked edges,
            for tree={
                grow'=0,
                draw,
                reversed=true,
                anchor=base west,
                parent anchor=east,
                child anchor=west,
                base=left,
                font=\large,
                rectangle,
                rounded corners,
                align=left,
                minimum width=4em,
                edge+={darkgray, line width=1pt},
                s sep=3pt,
                inner xsep=2pt,
                inner ysep=3pt,
                line width=0.8pt,
                ver/.style={rotate=90, child anchor=north, parent anchor=south, anchor=center},
            },
            where level=1{text width=11.5em,font=\normalsize,}{},
            where level=2{text width=18em,font=\normalsize,}{},
            where level=3{text width=8em,font=\normalsize,}{},
            where level=4{text width=5em,font=\normalsize,}{},
			[
			Speech-LLM Integration, ver
			    [
			      Text-based \textcolor{blueref}{(Sec.~\ref{sec:text-based-integration})}
			        [
			        Cascaded Integration \textcolor{blueref}{(Sec.~\ref{sec:cascaded-integration})}
    			        [
                             AudioGPT \cite{huang-2024-audio-gpt}{, } HuggingGPT \cite{shen-2023-hugginggpt}{, }\citet{dighe-2024-uncertainty}
                            , leaf, text width=46em
    			        ]
			        ]
			        [
		            LLM Rescoring \textcolor{blueref}{(Sec.~\ref{sec:llm-rescoring})}
			            [  
                            \citet{chen-2023-longform}{, }\citet{udagawa-2022-rescoring}{, }\citet{li-2023-asr-domain}{, }\citet{hu-2023-shallow-fusion}{, }\citet{huang-2024-fusion-comprehensive}
                            , leaf, text width=46em
			            ]
			        ]
                    [
                    LLM GER \textcolor{blueref}{(Sec.~\ref{sec:llm-ger})}
			            [  
                            HyPoradise \cite{chen-2023-hyporadise}{, }\citet{yang-2023-tap}{, }GenTranslate \cite{hu-etal-2024-gentranslate}{, }\citet{ma-2023-error-correction}{, }\\\citet{hu-2024-noise-robust}{, }Whispering Llama \cite{radhakrishnan2023whispering}{, }MMGER \cite{mu2024mmger}{, }\\\citet{yang2024large}{, }\citet{ko2024benchmarking}{, }LipGER~\cite{ghosh2024lipger}{, }\citet{li2024investigating}{, }\citet{tang2024contextualized} {, }\\ \citet{chen2024s} {, }\citet{dighe2024leveraging}
                            , leaf, text width=46em
			            ]
                    ]
			    ]
                [
                Latent-representation-based\\ \textcolor{blueref}{(Sec.~\ref{sec:latent-representation-based-integration})}
                    [
                    Convolutional Downsampling \textcolor{blueref}{(Sec.~\ref{sec:downsampling})}
                        [
                        \citet{hono-2023-integration}{, }\citet{yu-2024-connecting}{, }\citet{fathullah-2024-prompting}{, }LTU \cite{gong-2024-ltu}{, }\citet{pham-2024-comprehensive}{, }\\BLSP \cite{wang2023blsp}{, }SALM \cite{chen-2024-salm}{, }WavLLM \cite{hu-2024-wavllm}{, }DeSTA \cite{lu-2024-desta}{, }\\Llama-Omni \cite{fang-etal-2024-llamaomni}{, }Seed-ASR \cite{bai2024seed}{, }SpeechVerse \cite{das2024speechverse}{, }\\SLAM-ASR \cite{ma2024embarrassingly}{, }\citet{denisov2024teaching} {, }MaLa-ASR~\cite{yang2024mala}{, }\\WHISMA~\cite{li2024whisma}{, }\citet{lakomkin2024end}{, }\citet{noroozi2024instruction}{, }\citet{kang2024prompting}{, }\\ \citet{li2024using} {, }\citet{chen2024s}
                        , leaf, text width=46em
                        ]
                    ]
                    [
                    CTC Compression \textcolor{blueref}{(Sec.~\ref{sec:ctc-compression})}
                        [
                        \citet{hono-2023-integration}{, }Speech-Llama \cite{wu-2023-decoder-only}{, }\citet{ling-2024-fully-formatted}{, } \citet{fathullah2023towards}
                        , leaf, text width=46em
                        ]
                    ]
                    [
                    Q-Former \textcolor{blueref}{(Sec.~\ref{sec:qformer})}
                        [
                        \citet{yu-2024-connecting}{, }SALMONN \cite{tang-etal-2024-salmonn}{, }COSMIC \cite{pan-2023-cosmic}{, }XLLM \cite{chen-2023-xllm}{, }\\DeSTA \cite{lu-2024-desta}{, }Secap \cite{xu2024secap}{, }BLSP-KD \cite{wang2024blsp}{, } CoT-ST~\cite{du2024cot} 
                        , leaf, text width=46em
                        ]
                    ]
                    [
                    Other Adaptation Methods
                        [
                        SLM \cite{wang-2023-slm}{, } Qwen-Audio \cite{chu-2023-qwen-audio}{, } Qwen2-Audio \cite{chu-2024-qwen2-audio}{, } \\LauraGPT \cite{du-etal-2024-lauragpt}{, } LLaST \cite{chen2024llast}{, } BESTOW \cite{chen2024bestow}{, }\\ SALSA \cite{mittal2024salsa}{, } Ideal-LLM~\cite{xue2024ideal} {, }\citet{xu2024comparing}{, }\citet{chen2024s}
                        , leaf, text width=46em
                        ]
                    ]
                ]
			    [
		          Audio-token-based \textcolor{blueref}{(Sec.~\ref{sec:audio-token-based-integration})}
			        [
			        Semantic Tokens \textcolor{blueref}{(Sec.~\ref{sec:semantic-token})}
			            [
                            VoxtLM \cite{maiti-etal-2024-voxtlm}{, }SpeechGPT \cite{zhang-etal-2023-speechgpt}{, }TWIST \cite{hassid-etal-2023-twist}{, }\\Spirit-LM \cite{nguyen-etal-2025-spiritlm}{, } GPT-Talker \cite{liu-2024-gpt-talker}{, } BASE TTS \cite{lajszczak2024base}{, }\\Llama-Omni \cite{fang-etal-2024-llamaomni}{, }SpeechGPT-Gen \cite{zhang-etal-2024-speechgptgen}{, }  DiscreteSLU \cite{shon2024discreteslu}{, }\\ PSLM~\cite{mitsui2024pslm}{, }\citet{cheng2024task}{, }VSP-LLM~\cite{yeo2024visual}{, }\citet{xu2024comparing}
			            , leaf, text width=46em
			            ]
			        ]
			        [
                    Acoustic Tokens \textcolor{blueref}{(Sec.~\ref{sec:acoustic-token})}
			            [
                            \citet{hao-2024-boosting}{, } LauraGPT \cite{du-etal-2024-lauragpt}{, }\citet{neekhara2024improving}{, }Uniaudio~\cite{yang2023uniaudio}
			              , leaf, text width=46em
			            ]
			        ]
			        [
                    Semantic and Acoustic Tokens \textcolor{blueref}{(Sec.~\ref{sec:semantic-and-acoustic})}
			            [
                            AudioPaLM \cite{rubenstein-etal-2023-audiopalm}{, }Moshi \cite{defossez-etal-2024-moshi}{, }Hibiki~\cite{labiausse-etal-2025-hibiki}{, }\\ Emova~\cite{chen2024emova}{, }\citet{shen2024get}
			              , leaf, text width=46em
			            ]
			        ]
			    ]
			]
        \end{forest}
    }
    \caption{Taxonomy for speech-LLM integration. ``Latent-representation-based'' integration is categorized by different modality adaptation strategies. ``Other Adaptation Methods'' also includes studies that do not explicitly mention their modality adaptation methods. }
    \label{fig:taxonomy}
\end{figure*}

\section{Background} \label{sec:background}

\subsection{Language Modeling}

Language modeling dates back to statistical models like $N$-gram models~\citep{brown-etal-1992-ngram}, which were central to early NLP and automatic speech recognition (ASR) systems~\citep{bahl-etal-1983}.
Neural-network-based language modeling was introduced by~\citet{bengio-etal-2000}, formulating the probability
\[
 P(w_{1:T}) = \prod_{t=1}^T P(w_t | w_{1:t-1}),
\]
where $w_t$ is the $t$-th token (text subwords or speech tokens) and $w_{i:j} = (w_i, w_{i + 1}, \cdots, w_j)$ is the subsequence from $i$-th token to $j$-th token.

Advances in hardware enabled neural models to scale, leading to innovations such as sequence-to-sequence learning~\citep{sutskever-etal-2014-seq2seq} and the attention mechanism~\citep{bahdanau-etal-2015}.
The Transformer architecture~\citep{vaswani-etal-2017} revolutionalizes language modeling using self-attention by efficiently modeling long-range dependencies utilizing parallelized computation.
Decoder-only Transformer~\citep{liu-etal-2018} gains prominence through the multi-task learning paradigm with generative pre-training and discrinative fine-tuning~\citep{radford-etal-2018}.
Scaling these models~\citep{radford-etal-2019,brown-etal-2020} shows generalization to multiple tasks without explicit supervision and achieves comparative performance against task-specific models, which leads to the advent of LLM era.
Techniques such as instruction tuning~\citep{wei-etal-2022} and alignment to human preference via reinforcement learning~\citep{ouyang-etal-2022} further advance LLM capabilities.
For a comprehensive overview of LLMs, we refer the readers to LLM survey papers~\citep{zhao-etal-2023-survey-llm,minaee-etal-2024}.

\subsection{Speech Representations} \label{sec:background-speech-representations}

Speech is captured as waveform signals, sampled at a specific sampling rate and quantization value, and represented as a sequence of amplitude values.
For deep learning applications, they are often converted to speech representations with shorter sequence lengths, such as log Mel filterbanks, which we refer to as \textbf{filterbanks}.
Following the success of the unsupervised pre-training approach in other fields, self-supervised speech models (S3Ms)~\citep{baevski-etal-2020-wav2vec2,hsu-etal-2021-hubert,chen-etal-2022-wavlm} were introduced.
These models predict masked frames to learn representations that capture phonetic information \citep{choi-etal-2024}, along with a wide range of speech characteristics, such as speaker identity, paralinguistic information, and word-level information \citep{pasad-etal-2024}.
We refer to these contextualized frame-level features as \textbf{latent representations}.

To integrate speech (which is naturally continuous) with LLMs (which are designed to handle discrete tokens), \textbf{audio tokens} that are more discrete than previous representations have been studied recently.
There are two mainstream approaches to modeling speech as audio tokens:\footnote{Different studies use different terms for these tokens 
(e.g., \citet{lakhotia-etal-2021-gslm} used the term ``units'' for semantic tokens). 
We use the terms semantic tokens and acoustic tokens as in \citet{borsos-etal-2023-audiolm} and the term audio tokens to represent either or both of them similar to \citet{defossez-etal-2024-moshi}.} \textbf{semantic tokens}
and \textbf{acoustic tokens}.\footnote{We note that there are other efforts than semantic or acoustic tokens to discretize speech signals \citep{bai-etal-2024-dmel}.}
Semantic tokens can be obtained from S3Ms by discretizing latent representations using $k$-means.
Acoustic tokens have been studied in the line of audio codecs (e.g., MP3 \citep{iso11172-3}, Opus \cite{rfc6716}, etc.), which aim to compress audio signals.
Neural audio codecs \citep{zeghidour-etal-2021-soundstream,defossez-etal-2023-encodec,kumar-etal-2023-dac} have recently gained traction as powerful tools to improve coding efficiency as well as perceptual quality.
Because acoustic tokens exhibit better quality in producing output speech signals \cite{borsos-etal-2023-audiolm} but fail to capture semantic information like semantic tokens do, recent studies combine both to obtain better representations for speech-language modeling \citep{wang-etal-2023-viola,zhang-etal-2024-speechtokenizer,defossez-etal-2024-moshi}.

\section{Text-based Integration} \label{sec:text-based-integration}


This approach primarily utilizes text as both input and output for LLMs. 
One of the most direct implementations of this method is through \textbf{cascaded integration}, which is highly adaptable to various tasks.
Within tasks like ASR, LLMs are not limited to working with the final recognition results but can also operate on intermediate hypotheses. This allows for sophisticated operations like \textbf{LLM rescoring}, 
and \textbf{LLM GER} (Generative Error Correction).

\begin{figure}[t]
  \centering
  \includegraphics[width=\linewidth]{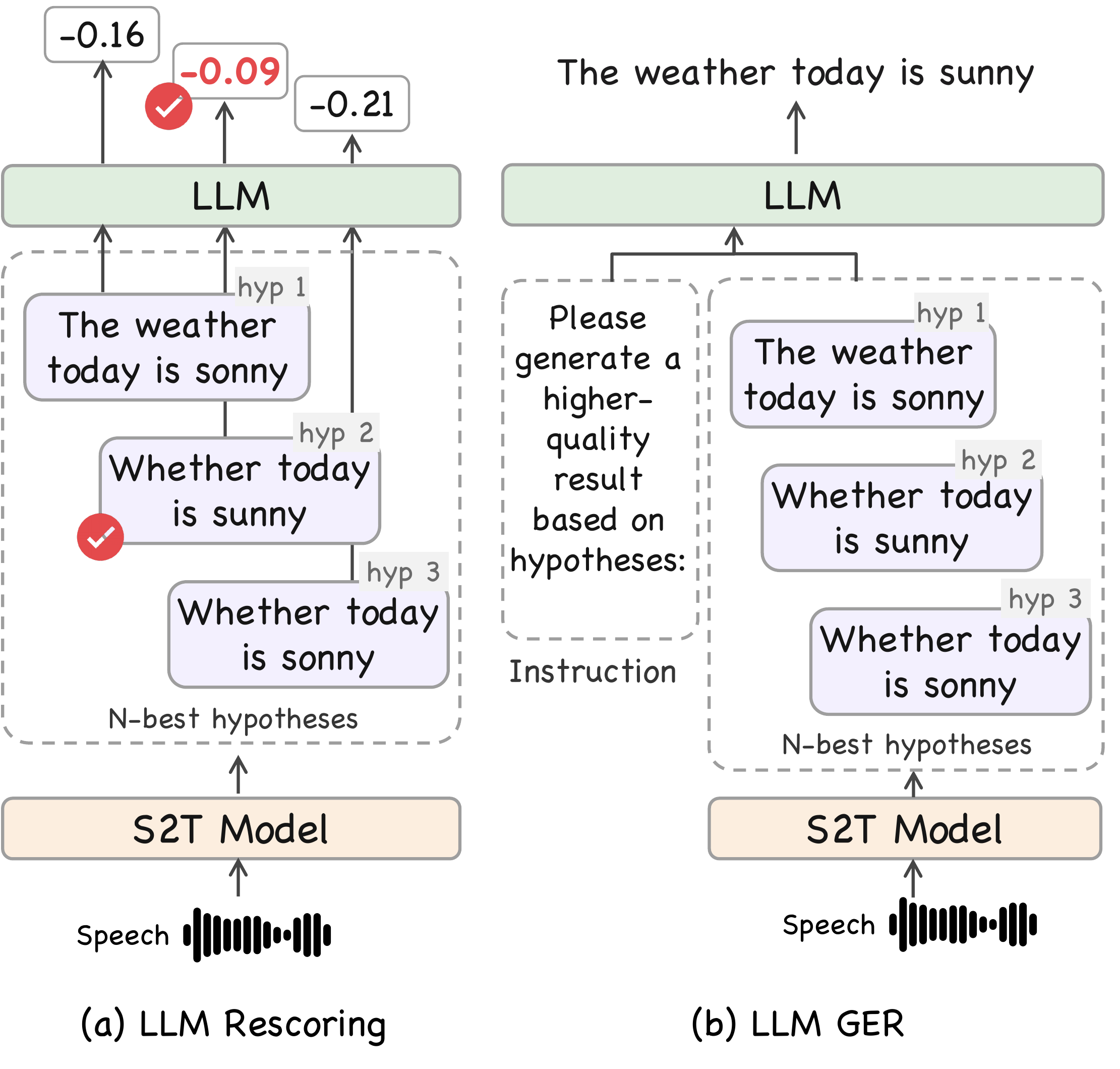}
  \caption{Two text-based approaches using hypotheses for improving speech-to-text tasks.}
  \label{fig:rescore}
\end{figure}

\subsection{Cascaded Integration} \label{sec:cascaded-integration}

Cascaded integration is the most simple and straightforward approach for integrating speech with LLMs.
This involves the backbone LLM being supported by ASR and Text-to-Speech (TTS) interfaces for handling speech inputs and outputs, as well as the LLM invoking external models to solve speech-related tasks.
This approach has been employed in models such as AudioGPT~\cite{huang-2024-audio-gpt} and HuggingGPT~\cite{shen-2023-hugginggpt} to expand the applications of LLMs. 
The primary advantage of cascaded integration is its ease of implementation, allowing various tasks to be integrated into a single model with minimal effort. 
However, this method suffers from accuracy problems due to error propagation and efficiency problems due to the latency of processing through multiple steps.

\subsection{LLM Rescoring} \label{sec:llm-rescoring}

Before research on LLMs explosively expanded, ASR models had already begun using rescoring mechanisms with external language models to improve transcription accuracy~\cite{mcdermott-2019-lm-fusion,shin-2019-sentence-scoring,salazar-2020-mlm-scoring,xu-2022-rescorebert}.
Recently, there have been efforts towards utilizing LLMs for rescoring, and it has been proven to be effective~\cite{chen-2023-longform,udagawa-2022-rescoring}. 

As shown in Figure \ref{fig:rescore} (a), this process involves generating a list of $n$-best hypotheses $\mathcal{H} = \{Y_1,Y_2,...,Y_n\}$  from the speech input $X$ through the initial ASR decoding, which is then re-evaluated using an LLM to find the hypothesis with higher linguistic coherence to improve accuracy.
The rescoring process can be expressed as
\begin{align*}
Y^* &= \text{argmax}_{Y_i \in \mathcal{H}} \ [(1\!-\!\lambda) \log p_{\text{AM}}(Y_i|X) \\
&\quad + \lambda \log p_{\text{LLM}}(Y_i)]
\end{align*}
where $p_{\text{AM}}$ and $p_{\text{LLM}}$ is the probability according to the ASR model (Acoustic Modle, AM) and the LLM, and $\lambda$ is a weight balancing the contributions of two models.


\subsection{LLM Generative Error Correction} \label{sec:llm-ger}

Based on the concept of rescoring, LLM GER represents a new approach that fully utilizes the generative capability of LLMs. 
As shown in Figure \ref{fig:rescore} (b), the method utilizes an instructional prompt together with the n-best hypotheses list to guide the LLM in generating transcription predictions. 
~\citet{chen-2023-hyporadise} described this task performed by the LLM as hypotheses-to-transcription (H2T), which they explore under both fine-tuning and few-shot settings.
Further studies have explored this direction with PEFT~\cite{hu-2024-noise-robust} or in-context learning~\cite{yang-2023-tap,ma-2023-error-correction}.

Instead of merely selecting the best hypothesis from a pre-existing set,
this method allows the LLM to generate a new transcription based on the hypotheses.
By doing so, LLMs can potentially produce a result with a quality better than all initial hypotheses. 
Such an approach has proven to be well-suited for other tasks such as speech-to-text translation (S2TT), which require more generative capabilities of the model~\cite{hu-etal-2024-gentranslate}.
\citet{lin-etal-2024-neko} adopted the mixture-of-experts architecture \citep{fedus-etal-2022,jiang-etal-2024-mixtral} to let different experts handle different types of generative errors produced by task-specific models including ASR and ST models.

\begin{figure*}[t]
  \includegraphics[width=\linewidth]{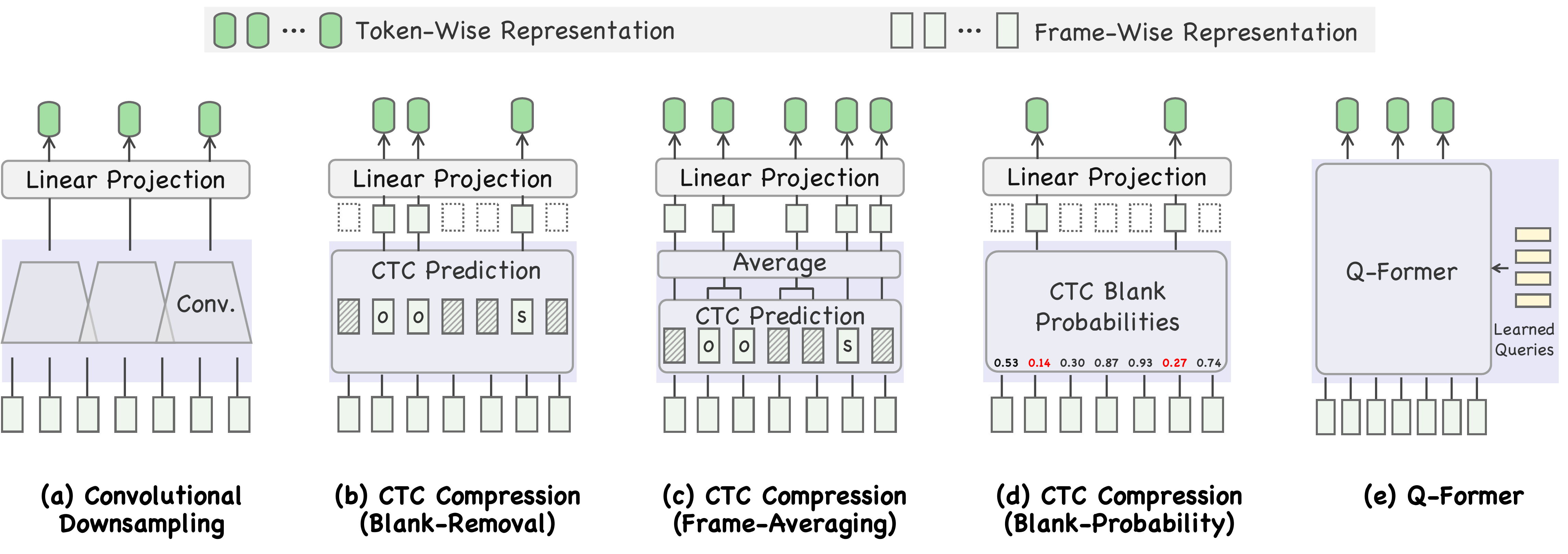}
  \caption{Different strategies for modality adaptation.}
  \label{fig:adapter}
\end{figure*}

\section{Latent-representation-based Integration} \label{sec:latent-representation-based-integration}

In this integration approach, 
a \textbf{speech encoder} is used to process the speech input and generate latent representations that are directly fed into the LLM, bypassing the embedding layer.
The speech encoder is usually a Transformer-based model, which can be pre-trained on large-scale speech data or trained from scratch for the speech--LLM integration. 

A critical issue of this approach is the sequence length gap between the speech and text modalities. Speech features, often sampled at rates of 50 to 100 frames per second, result in longer sequences compared to text tokens. Consequently, some form of \textbf{modality adaptation} mechanism is necessary to bridge these modalities, ensuring that the latent representations align with the LLM’s embedding space.

\subsection{Speech Encoder}
\label{sec:speech-encoder}

For the speech encoder, various pre-trained models can be utilized to provide representations learned from large-scale speech data.
This includes S3Ms (e.g. HuBERT~\cite{hsu-etal-2021-hubert}) as well as the encoder of pre-trained ASR models (e.g. Whisper~\cite{radford-etal-2023-whisper}). 
To connect the pre-trained speech encoder and the LLM, an \emph{adapter} (also referred to as a \emph{bridge network} or a \emph{module connector}) is usually adopted.

An alternative approach is to train a speech encoder from scratch, specifically for the LLM integration.
The structure of the speech encoder is usually a multi-layer Transformer~\cite{vaswani-etal-2017} encoder or Conformer~\cite{gulati-2020-conformer}.

\subsection{Modality Adaptation}
\label{sec:modality-adaptation}

The modality adaptation mechanism is adopted for the speech encoder to map original frame-wise representations to token-wise representations.  
This allows the LLM to process the speech sequence in a manner similar to how it processes the text sequence, without having to deal with overly long sequences caused by long-form speech inputs.

The adaptation is usually accomplished by the adapter between pre-trained models, which is introduced in \ref{sec:speech-encoder}, or any part of the self-designed trained-from-scratch speech encoder.
To this end, various adaptation methods have been proposed. Apart from some rarely-used strategies such as random downsampling~\cite{wang-2023-slm}, most of the methods can be categorized into three groups, as shown in Figure \ref{fig:adapter}: convolutional downsampling, CTC compression, and Q-former.
According to comparisons conducted by ~\citet{hono-2023-integration} and ~\citet{yu-2024-connecting}, the Q-former generally outperforms convolutional downsampling, which in turn outperforms CTC compression.
Next, we will detail these strategies in the following sections.

\subsubsection{Convolutional Downsampling} \label{sec:downsampling}

A basic strategy where the speech representation sequence is downsampled from its original length using convolutional layers~\cite{hono-2023-integration}.
When using the same number for kernel size and stride, it becomes equivalent to simply stacking multiple representation vectors together, which may only benefit parameter efficiency~\cite{fathullah-2024-prompting}.
For better alignment with the LLM's embeddings, 
some variants add a fully connected layer or a multi-head Transformer network after the convolutional layers~\citep{yu-2024-connecting}.


\subsubsection{CTC Compression} \label{sec:ctc-compression}

CTC compression involves two steps: (i) Training the speech encoder, or a part of it, on the ASR task based on Connectionist Temporal Classification (CTC)~\cite{graves-2013-ctc}; (ii) Compressing the representation sequence based on the results of CTC predictions. Several specific strategies for the step (ii) include: 

\paragraph{Blank-removal} 
Discarding all the frames predicted as blanks~\cite{hono-2023-integration,wu-2023-decoder-only}.

\paragraph{Frame-averaging} 
Averaging the latent representations of consecutive frames whose CTC predictions are the same label~\cite{hono-2023-integration,wu-2023-decoder-only}.

\paragraph{Blank-probability}
Discarding any frames with a CTC probability for blank exceeding a threshold~\cite{ling-2024-fully-formatted}.

\subsubsection{Q-Former} \label{sec:qformer}

Many studies~\cite{yu-2024-connecting,tang-etal-2024-salmonn,pan-2023-cosmic} adopt Q-Former for the modality adaptation. Q-Former is a Transformer-based module converting variable-length input sequences into fixed-length output query representations. It is initially proposed for vision--text modality alignment~\cite{li-2023-blip}.

\subsection{Training Strategy}
The optimal approach is to train the whole model consisting of the speech encoder and the LLM. However, fully fine-tuning the LLM is computationally expensive. Consequently, PEFT methods such as LoRA~\cite{xu-2024-lora}, are often adopted to reduce resource consumption.

In scenarios where a pre-trained speech encoder is utilized, the primary focus is training the adapter connecting two pre-trained models. Whether to fine-tune the speech encoder or the LLM with PEFT seems optional as the decision differs across studies.
On this topic, several studies~\cite{hono-2023-integration,pham-2024-comprehensive} have conducted systematic investigations by comparing the effects of fine-tuning versus freezing different modules.

As for training the speech encoder from scratch, it is commonly done jointly with fine-tuning LLM with PEFT. 
\citet{wu-2023-decoder-only} proposed employing a two-stage training process where PEFT is not initiated until the speech encoder has undergone some initial training, ensuring a more stable training progression.

\section{Audio-token-based Integration} \label{sec:audio-token-based-integration}

As explained in Section \ref{sec:background-speech-representations}, there are two types of audio tokens: semantic tokens and acoustic tokens.
Studies use either or both, treating them as independent tokens like text or converting them back to latent representations.
When semantic tokens are used for the output, a separate model to convert them into filterbanks and to waveform signals similar to TTS models is often used.

\subsection{Speech Language Models without LLMs}

\subsubsection{Semantic Token} \label{sec:semantic-token}
\citet{lakhotia-etal-2021-gslm} introduced the concept of generative spoken language modeling, which uses semantic tokens as both the input and output of a language model.
\citet{polyak-etal-2021} and \citet{kharitonov-etal-2022-pgslm} extended the idea by incorporating prosodic information.
These models only used audio, and although they could output temporarily coherent contents, semantic understanding ability, and audio quality remained as challenges.

\subsubsection{Acoustic Token} \label{sec:acoustic-token}
The use of acoustic tokens for language modeling was popularized in the task of TTS \citep{wang-etal-2023-valle,chen-etal-2024-valle2}, where a few seconds of sample speech is used to generate coherent speech as those samples given text.
\citet{wang-etal-2023-viola} extended the idea to cover multi-tasking of ASR, machine translation (MT), and TTS.

\subsubsection{Integration of Semantic and Acoustic tokens} \label{sec:semantic-and-acoustic}
AudioLM \citep{borsos-etal-2023-audiolm} was introduced to incorporate both semantic and acoustic tokens.
They adopted a two-stage approach, where the model first predicts semantic tokens and then subsequently predicts acoustic tokens.
\citet{zhang-etal-2024-speechtokenizer} employed distillation from semantic representation to the first quantizer to combine semantic and acoustic tokens.
Their generation approach is similar to \citet{wang-etal-2023-valle}, where an autoregressive model produces the first-layer token and a non-autoregressive model produces the rest.

\begin{table*}[t]
\centering
\resizebox{\linewidth}{!}{
\begin{tabular}{cccccc}
    \toprule
    \textbf{Task} & \makecell[c]{\textbf{Dataset}} & \textbf{Metrics} & \textbf{Model}  & \textbf{Integration Method} & \textbf{Performance} \\
    \midrule
    \multirow{22}{*}{ASR} & \multirow{12}{*}{\makecell[c]{\textbf{Librispeech}\\ \small \textit{dev-clean} | \textit{dev-other} | \\ \small \textit{test-clean} | \textit{test-other}}} & \multirow{12}{*}{WER $\downarrow$} &  Whisper-large-v2 \cite{radford-etal-2023-whisper} & Non-LLM  & --- | --- | 2.7 | 5.2 \\
    &&& HyPoradise \cite{chen-2023-hyporadise} & Text-based $\rightarrow$ LLM GER & --- | --- | 1.8 | 3.7 \\
    &&&  BLSP \cite{wang2023blsp} & Latent-representation-based $\rightarrow$ Convolutional Downsampling  & --- | --- | 10.4 | --- \\
    &&&  SpeechVerse \cite{das2024speechverse} & Latent-representation-based $\rightarrow$ Convolutional Downsampling  & --- | --- | 2.1 | 4.4 \\
    &&&  Seed-ASR \cite{bai2024seed} & Latent-representation-based $\rightarrow$ Convolutional Downsampling  & --- | --- | \textbf{1.5} | \textbf{2.8} \\
    &&&  SALMONN \cite{tang-etal-2024-salmonn} & Latent-representation-based $\rightarrow$ Q-Former  & --- | --- | 2.1 | 4.9\\
    &&&  SLM \cite{wang-2023-slm} & Latent-representation-based $\rightarrow$ Other Adaptation Methods  &   --- | --- | 2.6 | 5.0  \\
    &&& Qwen-Audio \cite{chu-2023-qwen-audio} & Latent-representation-based $\rightarrow$ Other Adaptation Methods   & 1.8 | 4.0 | 2.0 | 4.2 \\
    &&&  Qwen2-Audio \cite{chu-2024-qwen2-audio} & Latent-representation-based $\rightarrow$ Other Adaptation Methods  & \textbf{1.3} | \textbf{3.4} | 1.6 | 3.6\\
    &&& LauraGPT \cite{du-etal-2024-lauragpt} & Latent-representation-based $\rightarrow$ Other Adaptation Methods & --- | --- | 4.4 | 7.7 \\
    &&&  SpeechGPT-Gen \cite{zhang-etal-2024-speechgptgen} & Audio-token-based $\rightarrow$ Semantic Token & --- | --- | 2.4 | --- \\
    &&& SLAM-ASR \cite{ma2024embarrassingly} & Latent-representation-based $\rightarrow$ Convolutional Downsampling   & --- | --- | 1.9 | 3.8 \\
    &&& BESTOW \cite{chen2024bestow} & Latent-representation-based $\rightarrow$ Other Adaptation Methods   & --- | --- | --- | 3.2 \\
    &&& \citet{hao-2024-boosting} & Audio-token-based $\rightarrow$ Acoustic Tokens & 3.7 | 6.6 | 3.4 | 7.1 \\
    \cline{2-6}
    & \multirow{4}{*}{\makecell[c]{\textbf{Fleurs}\\ \small \textit{zh} | \textit{en}}} & \multirow{4}{*}{WER $\downarrow$} &  Whisper-large-v2 \cite{radford-etal-2023-whisper} & Non-LLM & 4.2 | 14.7 \\
    &&& \citet{huang-2024-fusion-comprehensive} w/ PaLM2 & Text-based $\rightarrow$ LLM Rescoring &  13.1 | --- \\
    &&&  Seed-ASR \cite{bai2024seed} & Latent-representation-based $\rightarrow$ Convolutional Downsampling &  \textbf{3.43} | ---  \\
    &&&  Qwen2-Audio \cite{chu-2024-qwen2-audio} & Latent-representation-based $\rightarrow$ Other Adaptation Methods  &   --- | \textbf{7.5}  \\
    &&&  DiscreteSLU \cite{shon2024discreteslu} & Audio-token-based $\rightarrow$ Semantic Token  &   12.6 | ---  \\
    \cline{2-6}
    & \multirow{4}{*}{\makecell[c]{\textbf{AISHELL-2} \\ \small \textit{Mic | iOS | Android}}} & \multirow{4}{*}{WER $\downarrow$} &  Seed-ASR \cite{bai2024seed} & Latent-representation-based $\rightarrow$ Convolutional Downsampling & \textbf{2.2} | \textbf{2.2} | \textbf{2.2} \\ 
    &&& Qwen-Audio \cite{chu-2023-qwen-audio} & Latent-representation-based $\rightarrow$ Other Adaptation Methods & 3.3 | 3.1 | 3.3 \\
    &&&  Qwen2-Audio \cite{chu-2024-qwen2-audio} & Latent-representation-based $\rightarrow$ Other Adaptation Methods & 3.0 | 3.0 | 2.9 \\
    &&& LauraGPT \cite{du-etal-2024-lauragpt} & Latent-representation-based $\rightarrow$ Other Adaptation Methods & --- | 3.2 | --- \\
    \cline{2-6}
    & \multirow{2}{*}{\makecell[c]{\textbf{VoxPopoli}\\ \small \textit{All}}} & \multirow{2}{*}{WER $\downarrow$} &  SpeechVerse \cite{das2024speechverse} & Latent-representation-based $\rightarrow$ Convolutional Downsampling  & \textbf{6.5} \\ 
    &&& AudioPaLM \cite{rubenstein-etal-2023-audiopalm} & Audio-token-based $\rightarrow$ Semantic and Acoustic Tokens & 9.8 \\
    \hline
    
    \multirow{11}{*}{S2TT} & \multirow{11}{*}{\makecell[c]{\textbf{CoVoST2}\\ \small \textit{en-de} | \textit{de-en} |\\ \small \textit{en-zh} | \textit{zh-en} |\\ \small \textit{es-en} | \textit{fr-en} |\\ \small \textit{it-en} | \textit{ja-en}}} & \multirow{11}{*}{BLEU $\uparrow$}
    &  Whisper-large-v2 \cite{radford-etal-2023-whisper} & Non-LLM & --- | 36.3 | --- | 18.0 | 40.1 | 36.4 | 30.9 | 26.1 \\
    &&&  GenTranslate \cite{hu-etal-2024-gentranslate} & Text-based $\rightarrow$ LLM GER  & --- | 39.2 | --- | 21.6 | 42.0 | 41.7 | --- | 22.9 \\
    &&&  GenTranslate-V2 \cite{hu-etal-2024-gentranslate} & Text-based $\rightarrow$ LLM GER & --- | 40.6 | --- | 23.3 | 43.6 | 42.7 | --- | 25.4 \\
    &&& BLSP \cite{wang2023blsp} & Latent-representation-based $\rightarrow$ Convolutional Downsampling & 24.4 | --- | 41.3 | --- | --- | --- | --- | --- \\
    &&&  Speech-Llama \cite{wu-2023-decoder-only} & Latent-representation-based $\rightarrow$ CTC Compression & --- | 27.1 | --- | 12.3 | 27.9 | 25.2 | 25.9 | 19.9 \\
    &&& SALMONN \cite{tang-etal-2024-salmonn} & Latent-representation-based $\rightarrow$ Q-Former & 18.6 | --- | 33.1 | --- | --- | --- | --- | --- \\
    &&&  Qwen-Audio \cite{chu-2023-qwen-audio} & Latent-representation-based $\rightarrow$ Other Adaptation Methods & 25.1 | 33.9 | 41.5 | 15.7 | 39.7 | 38.5 | 36.0 | --- \\
    &&& Qwen2-Audio \cite{chu-2024-qwen2-audio} & Latent-representation-based $\rightarrow$ Other Adaptation Methods & \textbf{29.9} | 35.2 | \textbf{45.2} | 24.4 | 40.0 | 38.5 | 36.3 | --- \\
    &&&  LauraGPT \cite{du-etal-2024-lauragpt} & Latent-representation-based $\rightarrow$ Other Adaptation Methods & --- | --- | 38.5 | --- | --- | --- | --- | --- \\
    &&& LLaST \cite{chen2024llast} & Latent-representation-based $\rightarrow$ Other Adaptation Methods & --- | 41.2 | --- | 24.8 | \textbf{46.1} | \textbf{45.1} | \textbf{43.0} | \textbf{28.8} \\
    &&&  AudioPaLM \cite{rubenstein-etal-2023-audiopalm} & Audio-token-based $\rightarrow$ Semantic and Acoustic Tokens & --- | \textbf{43.4} | --- | \textbf{25.5} | 44.2 | 44.8 | --- | 25.9 \\
    &&& Ideal-LLM \cite{xue2024ideal} & Latent-representation-based $\rightarrow$ Other Adaptation Methods & 25.9 | 38.5 | --- | --- | 41.5 | 40.0 | 38.0 | --- \\
    \hline
    
    \multirow{4}{*}{S2ST} & \multirow{4}{*}{\makecell[c]{\textbf{CVSS S2ST}\\ \small \textit{de-en} | \textit{zh-en} |\\ \small \textit{es-en} | \textit{fr-en} |\\ \small \textit{it-en} | \textit{ja-en}}} & \multirow{4}{*}{ASR-BLEU $\uparrow$} &  Translatotron 2 + pretraining &  &  \\
    &&&   + TTS aug \cite{jia-etal-2022-s2st} & \multirow{-2}{*}{Non-LLM} &  \multirow{-2}{*}{33.6 | 13.1 | 38.5 | 36.5 | 35.7 | 8.5} \\
    &&& & & \\
    &&& \multirow{-2}{*}{AudioPaLM \cite{rubenstein-etal-2023-audiopalm}} & \multirow{-2}{*}{Audio-token-based $\rightarrow$ Semantic and Acoustic Tokens} & \multirow{-2}{*}{\textbf{37.2} | \textbf{20.0} | \textbf{40.4} | \textbf{38.3} | \textbf{39.4} | \textbf{20.9}} \\
    \hline

    \multirow{18}{*}{TTS} & \multirow{3}{*}{\makecell[c]{\textbf{AISHELL-1}}} & \multirow{3}{*}{\makecell[c]{CER $\downarrow$ | SECS $\uparrow$ |\\ MOSNet $\uparrow$}} &  VALL-E Phone \cite{wang-etal-2023-valle} & Non-LLM & \textbf{4.75} | \textbf{0.91} | \textbf{3.22} \\
    &&& VALL-E Token \cite{wang-etal-2023-valle} & Non-LLM & 6.52 | \textbf{0.91} | 3.19 \\
    &&&  LauraGPT \cite{du-etal-2024-lauragpt} & Audio-token-based $\rightarrow$ Acoustic Tokens &  6.91 | 0.90 | 3.14 \\
    \cline{2-6}
    & \multirow{5}{*}{\makecell[c]{\textbf{LibriTTS}}} & \multirow{5}{*}{\makecell[c]{WER $\downarrow$ | SECS $\uparrow$ |\\ MOSNet $\uparrow$}} & VALL-E Phone \cite{wang-etal-2023-valle} & Non-LLM & \textbf{4.30} | 0.92 | 3.28 \\
    &&&  VALL-E Token \cite{wang-etal-2023-valle} & Non-LLM & 6.57 | \textbf{0.93} | 3.28 \\
    &&& SpeechGPT-Gen (zero-shot) \cite{zhang-etal-2024-speechgptgen} & Audio-token-based $\rightarrow$ Semantic Token & 3.10 | 0.63 | 3.63 \\
    &&&  VoxtLM \cite{maiti-etal-2024-voxtlm} & Audio-token-based $\rightarrow$ Semantic Token & --- | --- | \textbf{4.36} \\
    &&& LauraGPT \cite{du-etal-2024-lauragpt} & Audio-token-based $\rightarrow$ Acoustic Tokens &   8.62 | 0.91 | 3.26 \\
    \cline{2-6}
    & \multirow{5}{*}{\makecell[c]{\textbf{Topic-StoryCloze}}} & \multirow{5}{*}{\makecell[c]{TSC $\uparrow$}} &  Spirit-LM \cite{nguyen-etal-2025-spiritlm} & Audio-token-based $\rightarrow$ Semantic Token  & 82.9 \\
    &&& TWIST-1.3B \cite{hassid-etal-2023-twist} & Audio-token-based $\rightarrow$ Semantic Token & 70.6 \\
    &&&  TWIST-7B \cite{hassid-etal-2023-twist} & Audio-token-based $\rightarrow$ Semantic Token & 74.1 \\
    &&& TWIST-13B \cite{hassid-etal-2023-twist} & Audio-token-based $\rightarrow$ Semantic Token & 76.4 \\
    &&&  Moshi \cite{defossez-etal-2024-moshi} & Audio-token-based $\rightarrow$ Semantic and Acoustic Tokens & \textbf{83.6} \\
    \cline{2-6}
    & \multirow{5}{*}{\makecell[c]{\textbf{StoryCloze}}} & \multirow{5}{*}{\makecell[c]{SSC $\uparrow$}} & Spirit-LM \cite{nguyen-etal-2025-spiritlm} & Audio-token-based $\rightarrow$ Semantic Token & 61.0 \\
    &&&  TWIST-1.3B \cite{hassid-etal-2023-twist} & Audio-token-based $\rightarrow$ Semantic Token & 52.4 \\
    &&& TWIST-7B \cite{hassid-etal-2023-twist} & Audio-token-based $\rightarrow$ Semantic Token & 55.3 \\
    &&&  TWIST-13B \cite{hassid-etal-2023-twist} & Audio-token-based $\rightarrow$ Semantic Token & 55.4 \\
    &&& Moshi \cite{defossez-etal-2024-moshi} & Audio-token-based $\rightarrow$ Semantic and Acoustic Tokens & \textbf{62.7} \\
    
    \bottomrule
\end{tabular}
}
\caption{Quantitative comparison based on applications. The \textbf{Bold} denotes the best result.}
\label{tab:benchmarks}
\end{table*}

\subsection{Integration of LLMs into Speech Language Models}

The use of LLMs as an underlying model of speech language models has become more popular following the success of LLMs.
\citet{hassid-etal-2023-twist} showed that training speech language models using underlying pretrained text language models could improve performance.
VoxtLM \citep{maiti-etal-2024-voxtlm} uses OPT \citep{zhang-etal-2022-opt} as an underlying LLM and conducts multi-task fintuning it on ASR, TTS, and language modeling on text and semantic tokens.
AudioPaLM \citep{rubenstein-etal-2023-audiopalm} adopts a similar approach to \citet{borsos-etal-2023-audiolm} but also enables text input/output, leveraging PaLM \citep{chowdhery-etal-2023-palm} as the underlying LLM.
TWIST and SpeechGPT \cite{zhang-etal-2023-speechgpt} use Llama \cite{touvron-etal-2023-llama} as the underlying model and use semantic tokens as both input and output.
Spirit-LM \citep{nguyen-etal-2025-spiritlm} uses Llama 2 \citep{touvron-etal-2023-llama2} as the underlying model and uses a mixture of semantic and text tokens for both input and output.

\citet{defossez-etal-2024-moshi} introduced a new architecture for speech language modeling.
Similarly to previous studies, they trained a text LLM and a codec model similar to \citet{zhang-etal-2024-speechtokenizer}.
Their architecture enables duplex ability of listening and generating speech by using a hierarchical autoregressive model consisting of two Transformer modules, namely Temporal Transformer and Depth Transformer \citep{lee-etal-2022-rqtransformer}.
This architecture is also shown to be effective for speech-to-speech translation (S2ST)~\cite{labiausse-etal-2025-hibiki}.

\section{Comparative Analysis} \label{sec:comparative-analysis}


\subsection{Advantages and Disadvantages} \label{sec:proscons}

It is important to understand the advantages and disadvantages of the three integration approaches when employing them for different applications. We summarize their pros and cons as follows:

\begin{itemize}[left=0pt,itemsep=-2pt]
    \item \textbf{Degree of integration:} Latent-representation-based $>$ Audio-token-based $>$ Text-based.
    \item \textbf{Interpretability:} Text-based $>$ Audio-token-based $>$ Latent-representation-based.
    \item \textbf{Speech generation ability:} Text-based and audio-token-based approach can generate speech, whereas latent-representation-based approaches typically cannot.
\end{itemize}

Based on these characteristics, different approaches excel in different scenarios:
\begin{itemize}[left=0pt,itemsep=-2pt]
    \item Latent-representation-based and audio-token-based approaches are better than text-based approach, in scenarios where sufficient resources (data, computational power, and time) are available or when real-time processing is required. Deeper integration reduces error propagation and latency but demands more extensive resources.
    \item Text-based approach is better than latent-representation-based and audio-token-based approaches, in scenarios where resources are limited or when greater interpretability is required.
    \item Latent-representation-based approach is better than audio-token-based, in scenarios where speech is only considered as input. Latent-representation-based methods provide the deepest integration when generating textual or other downstream outputs.
    \item Audio-token-based is better than latent-representation-based, in scenarios where speech is also considered as output. Generating speech from latent representations remains challenging, making audio-token-based approaches more suitable in these cases.
\end{itemize}

\subsection{Quantitative Comparison} \label{sec:quantitative}

\revised{We provide a comprehensive performance comparison in Table~\ref{tab:benchmarks}, comparing studies with different integration approaches across multiple tasks: ASR, S2TT, S2ST, and TTS.
The table highlights key metrics such as word error rate (WER) and BLEU, alongside each model's integration method and backbone components.}

While Table~\ref{tab:benchmarks} summarizes many state-of-the-art results, direct comparisons can be challenging due to variations in backbone LLMs, acoustic front-ends, and training protocols. For instance, although BLEU scores on CoVoST~2~\cite{wang-etal-2021-covost2} indicate that deeper integration methods (e.g., latent-representation-based) often yield improved performance, model size and training resources also play a significant role. Future work should focus on more uniform experimental conditions to isolate the impact of each approach.

Nonetheless, the metrics in Table~\ref{tab:benchmarks} serve as a starting point for evaluating how well each method handles different tasks. The results suggest that deeper integration (e.g., latent-representation-based) can be beneficial when sufficient computational resources
and training data are available, whereas text-based methods often offer greater interpretability and simpler pipelines.

\subsection{\revised{Computational Trade-offs}}

\revised{As different methods have different computational requirements, it is difficult to have direct comparisons through a literature review. For numerical comparisons, comparing different approaches with a standard setting is required, which is one future direction. That said, we provide an overview of the computational requirements per category below:}

\revised{\begin{itemize}[left=0pt,itemsep=-2pt]
\item For text-based integration, the computational cost is equivalent to that of running separate models for ASR, LLM, and TTS. Rescoring based on $N$-best list increases the inference compute by a constant factor. For generative error correction, studies have explored full fine-tuning, parameter-efficient fine-tuning, and mixture-of-experts of LLMs. Parameter-efficient fine-tuning can provide better performance compared to full fine-tuning~\cite{chen-2023-hyporadise}. Mixture-of-experts requires high memory and computation for training, but shows good performance in handling multiple tasks with light inference.
\item For latent-representation-based integration, speech encoders are connected to LLMs through adapters. These adapters are lightweight, and studies have explored full fine-tuning and parameter-efficient fine-tuning, including freezing a part of the models. \citet{pham-2024-comprehensive} showed that applying LoRA to the LLM significantly enhances the performance, and for the encoder module, full fine-tuning yields the best performance, followed by partial fine-tuning, while partial fine-tuning is more cost-effective.
\item For audio-token-based integration, it involves audio token encoding/decoding and language modeling with audio tokens. For semantic tokens, feature extraction from S3Ms and training/assignment steps of k-means is required. They are lightweight compared to language modeling (the largest S3Ms have about 2B parameters), but efficient implementation is required for large-scale data. If the output is speech, a vocoder to convert semantic tokens to waveforms is required, which is also relatively lightweight compared to language modeling. For acoustic tokens, encoding and decoding are done using the same model, and the inference is lightweight. Computational cost for language modeling is similar to that of text-based ones.
\end{itemize}}

\revised{Overall, we can argue that the computational cost is bottlenecked by the size of the LLM that is used. Text-based integration and audio-token-based integration would require roughly the same computational bottleneck by LLMs, while latent-representation-based integration requires handling both the speech encoder and LLMs at the same time.}

\section{Challenges} \label{sec:challenges}



\subsection{Text-based Integration}

Text-based integration preserves the original LLM input modality, which is text. Because the model is already optimized for textual inputs, this approach typically requires the least adaptation from the LLM. However, transforming speech into text before feeding it into the LLM inevitably introduces a layer of abstraction and potential information loss such as prosody and emotion, which limits the downstream performance.

Therefore, the main challenge is how to convey the rich information of source speech through text. In current practice, the intermediate text is often generated based on the highest probability~\cite{chen-2023-hyporadise, yang-2023-tap, radhakrishnan2023whispering}. Although this strategy maximizes intermediate accuracy, it may not be optimal for the LLM and its final outputs. One possible direction is to inject controlled randomness during decoding to ensure additional diversity. However, degrading accuracy could also have a negative effect, creating a trade-off. How to balance diversity and accuracy remains an open research question that requires future investigation.

\subsection{Latent-representation-based Integration}

In contrast to text-based integration, latent-representation-based integration employs an intermediate representation \revised{closer to the source speech but more distant from the LLM’s native input space.} Consequently, the primary challenge lies in aligning these acoustic representations with the textual embedding space of the LLM. Section~\ref{sec:modality-adaptation} introduces multiple modality adaptation mechanisms to address this and enable LLMs to handle speech more effectively.

While the primary focus of most research introduced in this paper has been on adapting LLMs to \revised{handle speech data better}, there is comparatively less attention on how these adaptations affect performance on text-based tasks. The alignment of text and speech data, along with their latent representations, remains underexplored, with only a few studies focusing on this issue~\cite{wang2023blsp,lu-2024-desta}.

A promising solution for improving speech-text alignment is the generation of high-quality synthetic speech data to supplement real datasets, thus closing the gap between the speech modality and text-based LLMs. More thorough investigations into how to align speech and text representations in complex languages (beyond English) could also strengthen the multimodal capabilities of LLMs.

\subsection{Audio-token-based Integration}

As stated in Sections \ref{sec:background-speech-representations} and \ref{sec:audio-token-based-integration},
S3Ms (and their derived semantic tokens) mainly capture phonetic information~\cite{choi-etal-2024}, while acoustic tokens offer higher fidelity in generating speech signals~\cite{borsos-etal-2023-audiolm}.

Integrating both representations, as presented in Section \ref{sec:semantic-and-acoustic}, is one approach that requires further investigation.
While this approach achieves strong performance across various downstream tasks (Table~\ref{tab:benchmarks}), there are still limited studies on this approach, where the models are English-centric models that rely on millions of hours of speech data.
Minimizing computational requirements is essential to extend this approach to languages lacking the vast resources available for English.

Some studies suggest that latent-representation-based integration can outperform audio-token-based integration~\cite{wang-etal-2024-comparativestudydiscretespeech}. However, it is possible that this is due to the suboptimal representation of audio tokens~\cite{gat-etal-2023-discreteslm}, requiring further improvement of tokenization methods and comparison with other integration approaches.

\subsection{Fair Comparison Across Integration Approaches}

Beyond these integration-approach-specific challenges, there is a notable gap in comparing the different integration approaches under a unified setting.
Most existing works focus on one approach, making it difficult to assess their relative merits consistently.
A fair comparison among these integration methods could clarify how different factors affect performance. 
Developing standardized benchmarks, protocols, and reporting practices for speech--LLM research would help future work isolate the core differences between these approaches.

\subsection{Multilingualism}

\revised{
The challenge of multilingualism in speech--LLM integration is also significant. 
Most existing speech--LLM models are designed to handle only English, which limits their range of applicability. 
Although there are studies focusing on multilingual capabilities~\cite{hu-etal-2024-gentranslate,hu-2023-shallow-fusion,huang-2024-fusion-comprehensive}, the majority still rely on English-centric LLMs. 
Only a small fraction of research efforts~\cite{denisov-2024-multilingual} explicitly consider the multilingualism of pre-trained LLMs.
However, the effectiveness of a speech--LLM model is often constrained by the quality of the text-based backbone LLM. 
This dependency implies that any limitations in the backbone LLMs, especially in terms of handling multiple languages, can directly impact the performance of speech LLMs.  }

\subsection{Real-time Processing}

\revised{
Real-time processing presents another substantial challenge in the field of speech--LLM integration. 
Compared to text-based applications, speech-based applications such as ST more often require real-time processing to be practical in everyday use. 
However, the large size of LLMs typically leads to increased latency, which can neutralize some of the benefits gained from improved accuracy. 
While most speech LLMs are capable of handling tasks offline, there is a need for more innovative methods that can facilitate online processing. 
Deeper integration with latent representations or audio tokens tends to result in lower latency compared to text-based approaches. 
Recent models that focus on real-time applications, such as Moshi~\cite{defossez-etal-2024-moshi}, show promising starts in addressing this challenge. 
Developing such methods would significantly enhance the utility of LLMs in real-time speech applications, making them more viable for real-world usage.}

\section{Conclusion} \label{sec:conclusion}

In this survey, we systematically categorize the integration of LLMs with speech modality, outlining three primary approaches: text-based, latent-representation-based, and audio-token-based integration.
Each method demonstrates unique strengths in performing different speech-related tasks.
However, numerous challenges remain in this evolving field, presenting significant opportunities for future research.


\section*{Limitations}
In our survey of the integration of speech with LLMs, we have covered the significant developments within this field. 
Despite our comprehensive approach, we acknowledge that the rapid pace of development in speech processing and LLM means that some recent advances or discussions might have escaped our attention and are not fully addressed within this paper.

\revised{
\section*{Acknowledgment}
This work was supported by the SPRING program of Kyoto University and JSPS KAKENHI Grant Number JP23K28144 and JP23KJ1356.
}







\end{document}